\documentclass[runningheads]{llncs}
\usepackage[rgb,x11names,table]{xcolor}
\usepackage[T1]{fontenc}
\usepackage{amsmath}
\usepackage{amssymb}
\usepackage{subfig}
\usepackage{hyperref}
\hypersetup{colorlinks=true,citecolor=blue}
\usepackage{algorithm}
\usepackage{algorithmicx}
\usepackage{algpseudocode}
\usepackage{graphicx,verbatim}
\usepackage{mathrsfs}
\usepackage{bbding}
\usepackage{rotating}
\usepackage{multirow}
\usepackage{marginnote}
\usepackage{overpic}
\usepackage{colortbl}
\usepackage[labelfont=bf]{caption}
\usepackage{tabularx}
\usepackage{tikz}
\usepackage{float}
\usepackage{enumitem}
\usepackage{subcaption}
\usepackage{color}

\begin{document}
\title{Multimodal Federated Learning With Missing Modalities through Feature Imputation Network}

\author{Pranav Poudel \inst{2, 4} \and
Aavash Chhetri\inst{2} \and
Prashnna Gyawali\inst{3} \and
Georgios Leontidis \inst{1} \and
Binod Bhattarai \inst{1}$^{*}$
}
\authorrunning{P. Poudel et al.}
\institute{University of Aberdeen, UK\and NepAl Applied Mathematics and Informatics Institute for research, Nepal \and West Virginia University, USA \and Fogsphere (Redev.AI), UK \\}


\renewcommand{\thefootnote}{*}
\footnotetext[1]{Correspondence to: Binod Bhattarai <binod.bhattarai@abdn.ac.uk>}
\renewcommand{\thefootnote}{\arabic{footnote}} 
    
\maketitle              
    \begin{abstract}
Multimodal federated learning holds immense potential for collaboratively training models from multiple sources
without sharing raw data, addressing both data scarcity and privacy
concerns—two key challenges in healthcare. A major challenge in training multimodal federated models in healthcare is the presence of missing modalities due to multiple reasons, including variations in clinical practice, cost and accessibility constraints, retrospective data collection, privacy concerns, and occasional 
technical or human errors. Previous methods 
typically rely on publicly available real datasets or synthetic data to compensate for missing modalities. However, obtaining real datasets for every 
disease is impractical, and training generative models to synthesize missing modalities is computationally expensive and prone to errors due to the high 
dimensionality of medical data. In this paper, we propose a novel, lightweight, low-dimensional feature translator to reconstruct bottleneck 
features of the missing modalities. Our experiments on three different datasets (MIMIC-CXR,
NIH Open-I, and CheXpert), in both homogeneous and heterogeneous settings consistently improve the performance of competitive baselines. The code and implementation details are available at:\url{https://github.com/bhattarailab/FedFeatGen} 

\end{abstract}

\section{Introduction}

Multimodal learning has emerged as a transformative field of research within medical machine learning. By combining information from diverse sources like imaging, omics, and pathology, multimodal AI holds the potential to transform the healthcare landscape
\cite{acosta2022multimodal}. Inspired by the human capacity to integrate multisensory information for more effective perception and
interaction, these models leverage data from multiple modalities to construct a holistic representation of diseases, thereby significantly improving diagnostic accuracy \cite{venugopalan2021multimodal,shrestha2023medical}. However, effective training of such models typically requires substantial amounts of centralized data, which presents a major challenge, especially in healthcare settings, where data are often dispersed across multiple medical centers due to privacy concerns.

 \begin{figure}[!t]
    \centering
    \includegraphics[width=\textwidth]{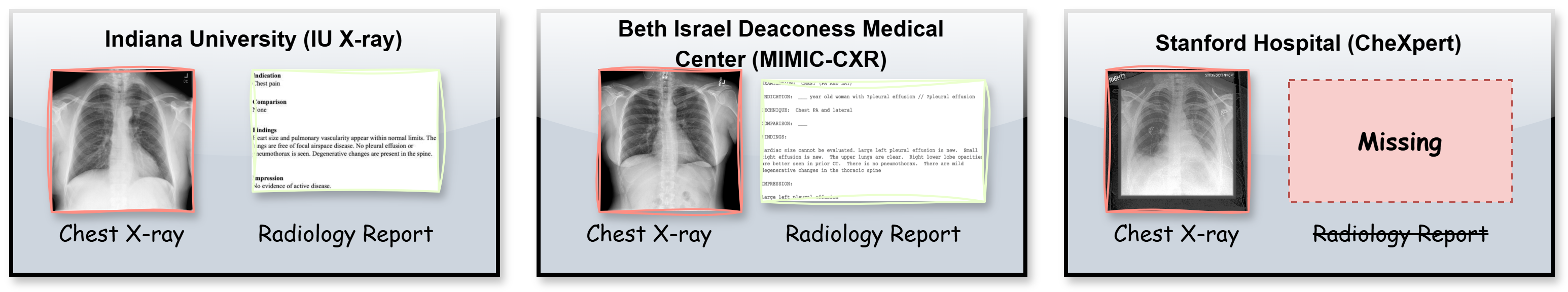}
    \caption{ 
    This figure shows sample of data from three different datasets collected at three different institutions. In CheXpert, there are only X-ray scans available, while the two other benchmarks have both X-ray scans and radiology reports. This demonstrates an instance of missing modality in a real-world scenario.}

    \label{fig:missing_modality}
\end{figure}

Multimodal federated learning is an important research topic due to its growing applications, and it offers an innovative approach for collaboratively training a shared model on heterogeneous healthcare data—such as X-rays, diagnostic reports, and time-series data—without sharing the raw data~\cite{thrasher2023multimodal,feng2023fedmultimodal}.
The main drawback of most prior work on multimodal federated learning is that it assumes the availability of complete modalities, overlooking the challenges posed by missing modalities~\cite{sachin2023multimodal,qayyum2022collaborative,chen2024medical}. In reality, as shown in Figure~\ref{fig:missing_modality}, multimodal systems often encounter challenges related to missing or incomplete modalities across medical centers due to the unavailability of clinical equipment, variability in data acquisition procedures, or limitations in data storage capacity ~\cite{thrasher2023multimodal}. Naïve strategies such as zero-imputation and uniform filling are often used to handle missing modalities~\cite{zheng2023autofed,le2024cross}. However, these approaches introduce bias into the global model, which adversely affects its performance and leads to suboptimal federated training. As a result, an increasing number of studies are aiming to fundamentally address the problem of missing modalities~\cite{wu2024deep,thrasher2023multimodal}. While research in the natural domain has been growing rapidly in recent years~\cite{yu2023multimodal,le2024cross,chen2022fedmsplit,sun2024towards}, there are still only a handful of studies in the medical domain~\cite{poudel2024car,saha2024examining}, partly due to the lack of standardized benchmarks.

The existing literature in multimodal federated learning are broadly categorized into three groups: public data-based methods~\cite{poudel2024car,yu2023multimodal}, class prototype-based methods~\cite{le2024cross}, and architecture-focused methods~\cite{chen2022fedmsplit,sun2024towards}. The primary drawback of public data-based approaches is that model performance heavily depends on the quality and representativeness of the available public data. Class prototype-based methods rely on impractical assumptions, such as having distinct and independent class prototypes, thus limiting their applicability to multiclass classification tasks only. Meanwhile, architecture-focused methods typically require specialized designs, restricting their generalizability. Recently, generative model-based approaches~\cite{saha2024examining} have also been explored, where missing reports are synthesized by pre-training generative models using data from available multimodal clients. Although this approach is conceptually simple and easy to implement, it demands significant data and computational resources for effective pre-training. Moreover, the data in healthcare are high-dimensional, such as Whole Slide Images (WSIs) have dimensions of 100,000, and reconstructing such high dimensional data incurs errors introducing un-necessary artifacts, which ultimately results in inferior performance~\cite{assran2023self}.

To address the limitations of existing methods, we propose learning low-dimensional bottleneck features of the missing modalities. To this end, we propose to train a lightweight feature-generation model in a federated-manner that conditionally synthesizes abstract, high-level modality-specific feature representations based on available input modalities (e.g., image features conditioned on text or vice versa). Our approach brings several advantages compared to the previous methods. First, our method does not need to have access of publicly available real data.
Also, the bottleneck features have low dimensionality, typically of a few hundred dimensions, which makes them easier to reconstruct with less error~\cite{assran2023self}. It is also lightweight, which is another advantage. To the best of our knowledge, this is the first work to compare the imputation of missing modalities at different level of input to offer a clear understanding of missing modality imputation strategies in federated scenarios where modality completeness cannot be guaranteed. 

To validate our idea, we conducted experiments simulating real-world federated learning scenarios using standard medical imaging datasets (MIMIC-CXR~\cite{johnson2019mimic}, NIH Open-I~\cite{demner2016preparing}, and CheXpert~\cite{irvin2019chexpert}). These experiments evaluate performance across configurations with varying ratios of multimodal and unimodal clients (image-only or text-only), reflecting typical data heterogeneity. Our experimental results outperform na\"ive baselines as well as the most closely related generative method, and achieve performance competitive with methods based on publicly available datasets.

Our key contributions are:

\begin{itemize}
    
    \item We propose a feature imputation method trained to handle missing modalities in multimodal federated learning. 
    \item  We present the first direct comparison and extensive evaluations of imputation strategies operating at the feature level versus the input (raw data) level in the context of federated learning.
    
\end{itemize}
\section{Method}
\subsection{Problem Formulation}
     We consider a multimodal federated learning setting with $C$ clients, each having its private dataset $D_C$ with $n_C$ samples. The $i^{th}$ data sample in $D_C$ is represented by tuple $(\{X_m^{(i)}\}_{m=1}^{M_C}, Y^{(i)})$, where $Y^{(i)}$ and $M_C$ represent the label set and the number of modalities in the $C^{th}$ client respectively. Without loss of generalizability, we assume a scenario with two modalities: image $(I)$ and text$(T)$. All clients utilize the same global architecture, which consists of modality-specific encoders $(f_e)$, concatenation-based fusion $(\oplus)$, and a classifier head $(f_c)$. Hence, we define the complete model as the set $\{f_e^{I}, f_e^{T},\oplus, f_c \}$.
\begin{equation}
    \underset{w}{\operatorname{argmin}}~L(\boldsymbol{w}) =  \sum_{c = 1}^{C} \frac{|D_c|}{|D|}  \text{  } l_c(\boldsymbol{w}, D_c)
    \label{eq: global_optimization}
\end{equation}
\begin{equation}
    \text{where, }l_c(\boldsymbol{w}, D_c) = \frac{1}{|D_c|} \sum_{(X_I^{(i)}, X_T^{(i)}, Y^{(i)}) \in D_c} \hspace{-0.75cm} \mathcal{L}(f_c(f_e^I(X_I^{(i)}) \oplus f_e^T(X_T^{(i)})), Y^{(i)})
    \label{eq: local_optimization}
\end{equation}

\noindent Here, \( D \) represents the combined dataset across all clients, while \( L\) denotes the local loss function at the client level for the model \( \{ f_e^I, f_e^T, f_c \} \) when applied to a data sample \( (X_I^{(i)}, X_T^{(i)}, Y^{(i)}) \). A widely used method for handling missing modalities is imputation sampled from a distribution. When a text sample is absent, the local loss function becomes:

\begin{equation}
\mathcal{L}(f_c(f_e^I(X_I^{(i)}) \oplus \psi), Y^{(i)})
\quad \text{where} \quad 
\psi \sim 
\begin{cases}
0 & \text{if zero-filling} \\
\mathcal{U}(0, 1) & \text{if uniform sampling}
\end{cases}
\label{eq: local_optimization_naive}
\end{equation}

\noindent Naively optimizing Equation \ref{eq: local_optimization_naive} results in sub-optimal performance.

\begin{figure}[!t]
    \centering
    \includegraphics[width=\textwidth]{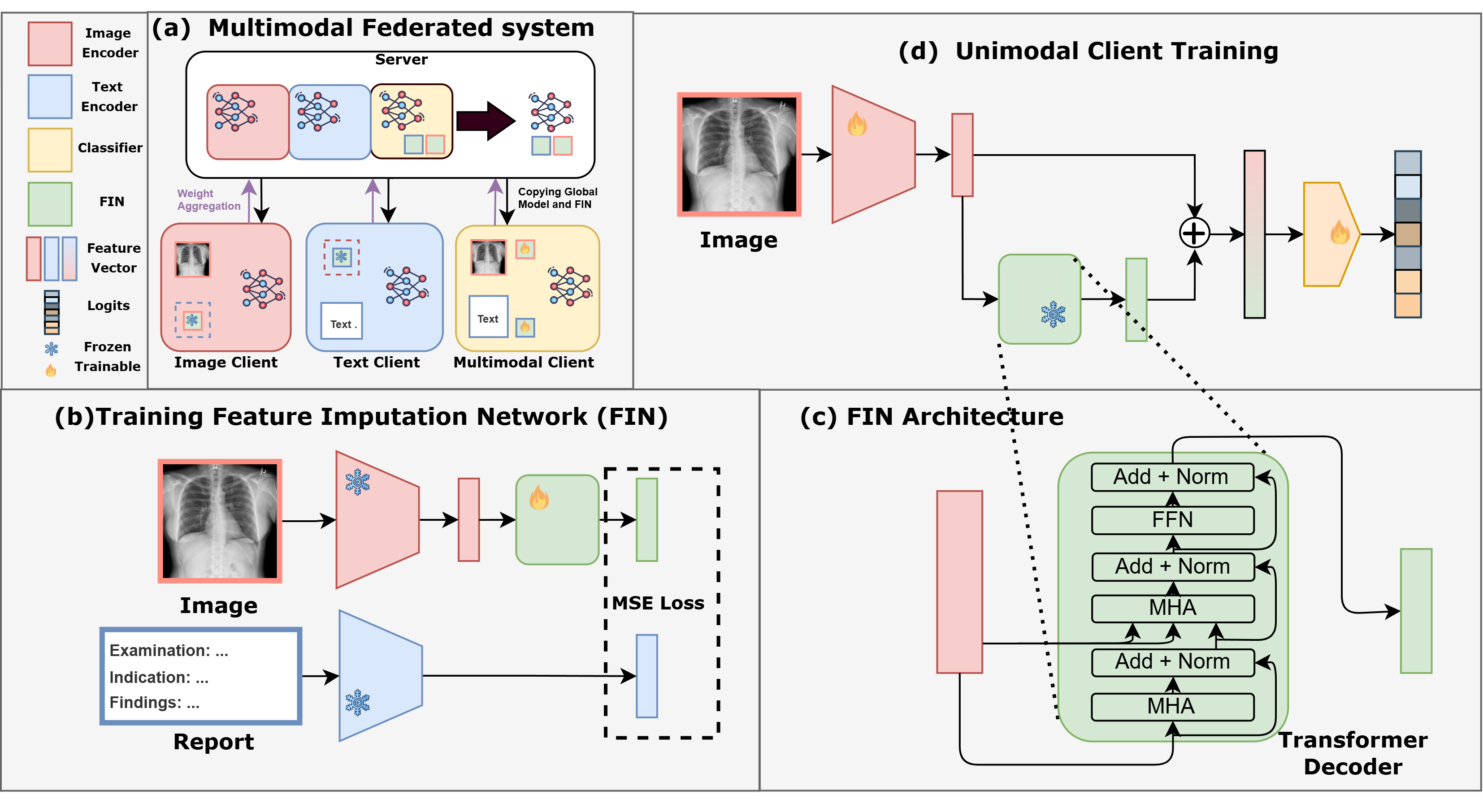}
    \caption{Illustration of \textbf{Feature Imputation Network}-based Multimodal Federated Learning. \textbf{(a)} Multimodal Federated Learning system with different types of clients. \textbf{(b)} Training of the Feature Imputation Network in multimodal client. \textbf{(c)} Architecture of Feature Imputation Network  \textbf{(d)} Unimodal image client training with the help of the Feature Imputation Network.}

    \label{fig:method}
\end{figure}

To mitigate such issues, we introduce the Feature Imputation Network(FIN), which approximates the missing modality feature based on the available modality. For modality-incomplete samples in unimodal clients, this network reconstructs the bottleneck features of the missing modality conditioned on the available one. Our overall method is illustrated in Figure~\ref{fig:method}.

\subsection{Feature Imputation Network (FIN)}
The feature imputation network generates the feature vector of the missing modality from the available one. 
Let $z_I^{(i)} = f_e^I(X_I^{(i)})$ denote the latent feature representation extracted by the image encoder for the $i^{th}$ sample, and $z_T^{(i)} = f_e^T(X_T^{(i)})$ denote the latent text feature representation.
We define imputation networks $\Phi_T$ and $\Phi_I$ such that $\Phi_T$ aims to approximate the text features from image features ($\Phi_T: z_I \mapsto \hat{z}_T$, where $\hat{z}_T \approx z_T$) and $\Phi_I$ aims to approximate image features from text features ($\Phi_I: z_T \mapsto \hat{z}_I$, where $\hat{z}_I \approx z_I$). At the start of each communication round, the server dispatches imputation networks $(\Phi_T, \Phi_I)$ and a complete global model $\{f_e^{I}, f_e^{T},\oplus, f_c \}$ to all clients. 
At multimodal clients, after training the model on local data for $k$ steps, we proceed to train the imputation networks. Without loss of generalization, let us assume we are training $\Phi_T$ to approximate text features. We first generate a pool of paired image-text feature vectors $P_c = \{(z_I^{(i)}, z_T^{(i)}) \mid (X_I^{(i)}, X_T^{(i)}, \cdot) \in D_c\}$.  As shown in Figure~\ref{fig:method} (b),  this pool $P_c$ is then used to train $\Phi_T$ by minimizing the Mean Squared Error (MSE) between the predicted text feature $\hat{z}_T^{(i)} = \Phi_T(z_I^{(i)})$ and the ground truth text feature $z_T^{(i)}$:

\begin{equation}
 \mathcal{L}(\Phi_T) = \frac{1}{|D_c|} \sum_{i=1}^{|D_c|} \| \Phi_T(z_I^{(i)}) - z_T^{(i)} \|^2_2
 \label{eq: Text_feature_imputation_optimization}
\end{equation}

After optimizing $\Phi_T$ (and symmetrically $\Phi_I$) for $k$ steps, both the updated main model and the updated imputation networks are uploaded to the server for aggregation. To complete the missing modalities,  the unimodal clients receive both the model and the imputation network at the start of each round, and use the imputation network (e.g., $\Phi_T$) purely for inference to generate features for missing modalities as shown in Figure~\ref{fig:method}(d).
The inferred features are concatenated with the bottleneck feature of the counterpart available modality and fed into the task-specific learnable network.  And, the objectives for the multimodal learning in the clients with only image and text become Equations~\ref{eq: local_optimization_imputation} and~\ref{eq: local_optimization_imputation_i}, respectively.

\begin{equation}
\mathcal{L}(f_c(f_e^I(X_I^{(i)}) \oplus \textcolor{blue}{\Phi_T}(f_e^I(X_I^{(i)})), Y^{(i)})
\label{eq: local_optimization_imputation}
\end{equation}

\begin{equation}
\mathcal{L}(f_c(f_e^T(X_T^{(i)}) \oplus \textcolor{blue}{\Phi_I}(f_e^T(X_T^{(i)})), Y^{(i)})
\label{eq: local_optimization_imputation_i}
\end{equation}

After optimizing for $k$ steps, the model is uploaded to the server, where models from all clients and feature imputation networks from multimodal clients are aggregated. We have implemented a simple 6-layer Transformer decoder~\cite{vaswani2017attention} with n=4 heads and 1024 feed-forward dimensions as our feature imputation network. This choice is motivated by the demonstrated success of Transformers in numerous cross-modal tasks. We employ FedAvg~\cite{mcmahan2017communication} as the aggregation strategy in the server.
\section{Experiments and Results}
\subsection{Datasets and Setups:}
Following \cite{poudel2024car}, we utilize three publicly available datasets- MIMIC-CXR \cite{johnson2019mimic}, NIH Open-I \cite{demner2016preparing}, and CheXpert \cite{irvin2019chexpert}—to design two experimental setups: Homogeneous and Heterogeneous.
Both setups consist of frontal chest X-rays and common validation and test sets derived from the official MIMIC-CXR splits, where the global model is validated and tested. In the homogeneous setup, 10 clients each contain training data from 810 patients, all sampled from MIMIC-CXR. Conversely, the heterogeneous setup consists of eightimage-only clients, each with data from 900 patients in CheXpert, and two multimodal clients, each containing data from 1,116 patients in NIH Open-I. This setup reflects real-world situations where data characteristics and label distributions vary across clients types. For simplicity, we denote client configurations using the format \textbf{I:T:M}, where I, T, and M represent the number of image-only, text-only clients, and the number of multimodal clients, respectively.

\subsection{Implementation Details:} 
We employ pre-trained ResNet-50 \cite{he2016deep} and BERT-base \cite{devlin2018bert} as the image and text encoders, respectively. Their outputs are transformed into 256-dimensional vectors and L2-normalized before fusion. We use a straightforward concatenation approach for multimodal fusion, followed by a linear layer for classification. The models are trained locally using Adam \cite{kingma2014adam} with a learning rate of $1e^{-4}$ for three epochs per communication round, totaling 30 rounds. For evaluation, we measure macro AUC (the average area under the Receiver Operating Characteristic curve) on the multimodal test set, following the evaluation protocol outlined in \cite{poudel2024car}. The reported values represent the mean results from experiments conducted with three random seeds.

\subsection{Baselines:}We compare our method against two na\"ive but common approaches: Zero-filling and Uniform-filling, along with state-of-the-art methods that depend on generative models and public datasets. Zero-filling imputes the missing data using zero vectors, while Uniform-filling imputes the missing data with feature vectors sampled from uniform distributions.

\noindent \textbf{R2Gen}~\cite{chen-emnlp-2020-r2gen} is a generative model that generates radiology reports from X-ray images. We trained R2Gen in a federated manner and subsequently used it to generate missing reports from images. For a fair comparison, the original ResNet-101 visual feature extractor was replaced with ResNet-50 to maintain consistency with our feature imputation approach.

\noindent \textbf{CAR-MFL}~\cite{poudel2024car} It is a state-of-the-art method in multimodal federated learning with missing modalities. This method is not directly comparable to ours, as it relies on publicly available real datasets to fill the missing modality gap.

\begin{table}[!ht]
\centering
\setlength\tabcolsep{3pt}
\renewcommand{\arraystretch}{1.30}
\caption{AUC$\uparrow$ Performance in Homogeneous and Heterogeneous setups in Unimodal Image Client Settings}
\begin{tabular}{ccccc}
\hline
\textbf{Partition}        & \multicolumn{3}{c}{\textbf{Homogeneous}}          & \multicolumn{1}{l}{\textbf{Heteregeneous}} \\ \hline
\multicolumn{1}{c}{I:T:M} & 8:0:2           & 6:0:4          & 4:0:6          & 8:0:2                                      \\ \hline
Zero-fillings              & 79.8            & 82.81          & 86.94         & 72.76                                      \\
Uniform-fillings           & 80.6            & 84.83          & 87.79          & 71.16                                      \\
R2Gen          & 77.32           & 83.1          & 86.83          & 67.32                                      \\
Feature Imputation (Ours)       & \textbf{86.16} & \textbf{87.61} & \textbf{89.31} & \textbf{77.94}                             \\ \hline
\end{tabular}
\label{tab:perform_1}
\end{table}

\begin{table}[h]
\centering
\setlength\tabcolsep{3pt}
\renewcommand{\arraystretch}{1.30}
\caption{AUC$\uparrow$ Performance in Homogeneous Across Various Client Settings. An asterisk (*) indicates that the method is not directly comparable to other methods.
}
\begin{tabular}{ccccccc}
\hline
Partition          & \multicolumn{6}{c}{Homogeneous}                                                                    \\ \hline
I:T:M              & 0:8:2          & 0:6:4         & 0:4:6          & 4:4:2          & 3:3:4          & 2:2:6          \\ \hline
\rowcolor{gray!20}
CAR-MFL* & 89.22          & 90.12             &  90.06              & 88.93       & 89.62              & 89.94              \\
Zero-fillings       & 86.18          & 86.8             & 88.21              & 80.3              & 84.57              & 87.09              \\
Uniform-fillings    & 88.37          & 88.64         & 89.27          & 85.74          & 86.9           & 88.49          \\
Feature Imputation (Ours) & \textbf{88.98} & \textbf{89.3} & \textbf{89.52} & \textbf{88.79} & \textbf{88.12} & \textbf{89.12} \\ \hline
\end{tabular}
\label{tab:perform_2}
\end{table}

\subsection{Quantitative Results}

Tables \ref{tab:perform_1} and \ref{tab:perform_2} present the performance of all baseline methods and our proposed approach across various client configurations in both experimental setups. As shown in Table \ref{tab:perform_1}, Feature Imputation method significantly outperforms all other imputation techniques in both homogeneous and heterogeneous settings, demonstrating its applicability to real-world, challenging scenarios. Remarkably, with only two multimodal clients (8:0:2), Feature Imputation achieves performance comparable to that of zero-filling in a six-client multimodal setting (4:0:6). This demonstrates that our method has a tremendous advantage even when only a few clients have data with complete modalities. Whereas, our competitor method severely fails to generalize in such a scenario as demonstrated by the performance gap of nearly $10\%$.
Similarly, in Table \ref{tab:perform_2}, which details performance across various homogeneous client settings, our Feature Imputation method consistently surpasses the other baseline imputation techniques (Zero-fillings and Uniform-fillings). Notably, our approach achieves results highly competitive with the CAR-MFL method, even though CAR-MFL benefits from access to public datasets to fill modality gaps, a condition our method does not require, highlighting its practical applicability.

Predicting in the representation space rather than the input space simplifies the task and encourages the model to learn abstract, high-level, meaningful features \cite{assran2023self}. Generative models make predictions in token space, which are prone to generating unnecessary or irrelevant tokens—especially when the model is trained on limited data. Consequently, both the feature extractor and the classification head are affected by noisy gradient updates, leading to degraded performance. In contrast, feature imputation methods operate directly in the representation space, reducing the likelihood of injecting irrelevant information. Moreover, these methods influence only the classification head during training, leaving the learned features of the missing modality encoders unaffected by gradient updates. This explains why even simple strategies like zero-filling or uniform-filling sometimes outperform generative models.

\subsection{Qualitative Results}
\begin{figure}[h]
    \centering
    \includegraphics[width=\textwidth]{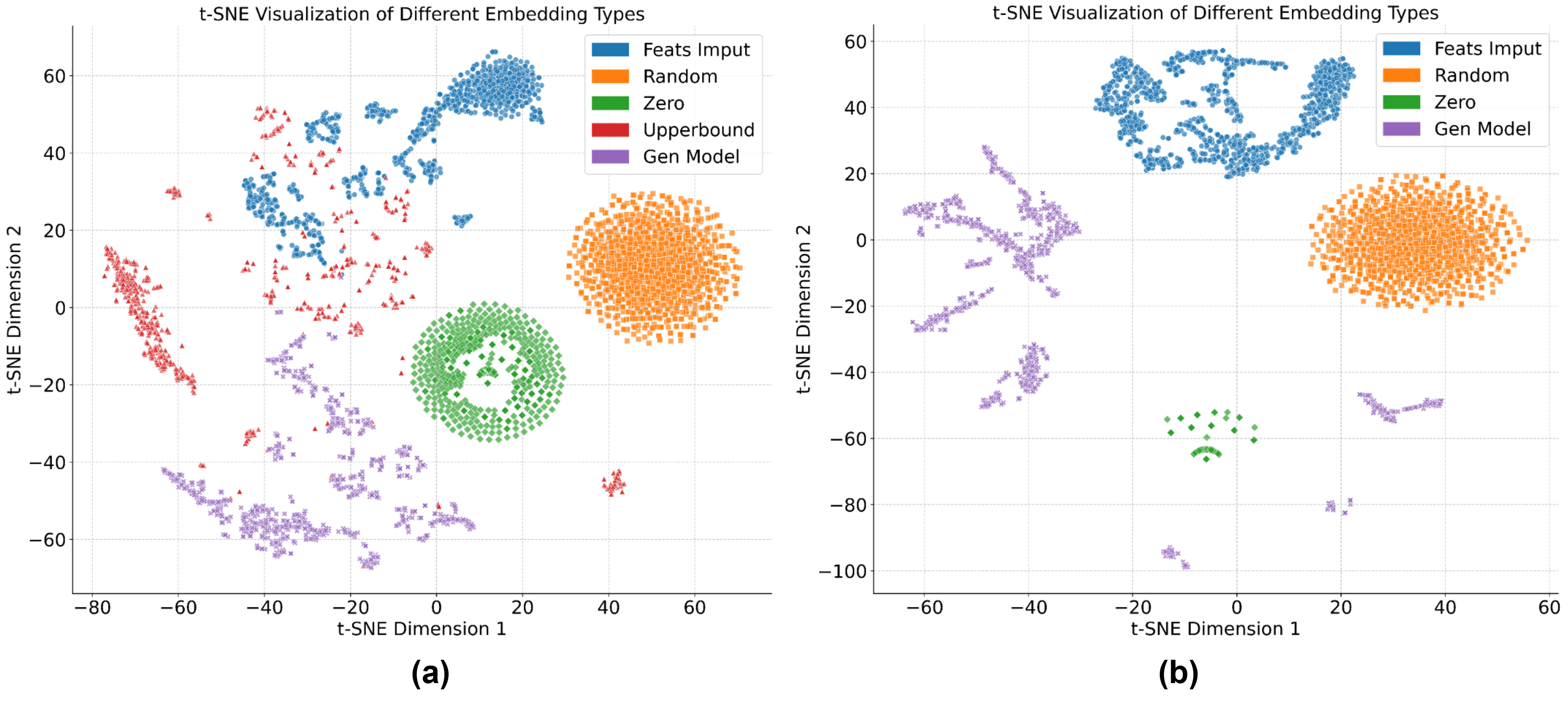}
    \caption{t-SNE plot of feature vectors from the model trained in \textbf{(a)} the homogeneous setup and \textbf{(b)} the heterogeneous setup. In the Figure,  \texttt{Upperbound} refers to the model trained in a federated manner with complete modalities. Feature vectors are generated using the validation data.}
    \label{fig:tsne_plot}
\end{figure}

Figure \ref{fig:tsne_plot} displays a t-SNE visualization comparing different feature representations derived from the validation dataset, specifically within the 8:0:2 federated setting (8 image-only, 2 multimodal clients). This allows for a qualitative assessment of the embeddings generated by our Feature Imputation network (approximating text features from image features) against those resulting from baseline imputation techniques (zero/uniform filling, text encoder output of model trained in generated text report) and an ideal upper-bound model. The \texttt{upper-bound} is the scenario when all the clients have complete modalities.
In the homogeneous setup, the representations generated by \texttt{upper-bound} model (represented by \textcolor{red}{red} dots) and our Feature Imputation network (represented by \textcolor{blue}{blue} dots) exhibit notably similar structural characteristics, primarily forming one large cluster accompanied by smaller scattered clusters, with some degree of overlap between them. In contrast, embeddings from the generative model trained on raw generated data (\textcolor{purple}{purple}) do not closely align with the structure of the upper-bound representations. Additionally, the zero-imputation (\textcolor{green}{green}) and uniform vector imputation methods (\textcolor{orange}{orange}), which produce embeddings from fixed distributions, result in dense clusters localized in single, confined regions.

\subsection{Computational Complexity and Communication Cost}
\begin{table}[!htbp]
\centering
\setlength\tabcolsep{3pt}
\renewcommand{\arraystretch}{1.30}
\caption{Comparison of model components in terms of approximate parameter counts and FLOPs (Floating Point Operations Per Second). FLOPs were estimated using the \texttt{fvcore } library.}

\begin{tabular}{ccc}
\hline
\textbf{Model}             & \textbf{Parameters} & \textbf{FLOPS} \\ \hline
Generative Model (R2Gen)   & 59.74 M          & 94.059 G \\
Feature Imputation Network & 6.324 M           & 6.318 M  \\ \hline
\end{tabular}
\label{tab:compute}

\end{table}

Table~\ref{tab:compute} presents a comparison between the Feature Imputation Network and the Generative Model in terms of computational efficiency.  We can see that the Feature Imputation Network reduces communication cost by nearly 10× per round and computational costs by approximately 1000× per inference.

\section{Conclusion}
We presented a feature imputation network designed to synthesize feature vectors for missing modalities using available data within a multimodal federated learning framework. Extensive experiments demonstrated that our approach significantly outperforms common baselines, including input-level generative models. Future research could explore extensions to more complex multimodal scenarios and investigate alternative architectures for the feature imputation network.

%
%
%
\bibliographystyle{splncs04}
\bibliography{PaperID-1085}

\end{document}